\documentclass[11pt]{article}
\usepackage{coling2020}
\usepackage{times}
\usepackage{url}
\usepackage{latexsym}
\usepackage{array}
\usepackage{amsfonts}
\usepackage{multirow}
\usepackage{amsmath}
\usepackage{graphicx}
\graphicspath{ {/} }
\usepackage{amssymb}
\usepackage{caption}
\usepackage{subfigure}

\colingfinalcopy 

\title{Hierarchical Bi-Directional Self-Attention Networks for Paper Review Rating Recommendation}

\author{Zhongfen Deng\textsuperscript{1}, Hao Peng\textsuperscript{2, 3}, Congying Xia\textsuperscript{1} , Jianxin Li\textsuperscript{2}, Lifang He\textsuperscript{4} , Philip S. Yu\textsuperscript{1}  \\
  \textsuperscript{1}Department of Computer Science, University of Illinois at Chicago, Chicago, IL, USA \\
  \textsuperscript{2}Beijing Advanced Innovation Center for Big Data and Brain Computing, Beihang University, Beijing, China \\
  \textsuperscript{3}School of Cyber Science and Technology, Beihang University, Beijing, China\\
  \textsuperscript{4}Department of Computer Science and Engineering, Lehigh University, Bethlehem, PA, USA \\
  {\tt \{zdeng21,cxia8,psyu\}@uic.edu, \{penghao,lijx\}@act.buaa.edu.cn} \\
  {\tt lih319@lehigh.edu} \\}

\date{}

\def\methodname{HabNet}
\begin{document}
\maketitle
\begin{abstract}
Review rating prediction of text reviews is a rapidly growing technology with a wide range of applications in natural language processing. 
However, most existing methods either use hand-crafted features or learn features using deep learning with simple text corpus as input for review rating prediction, ignoring the hierarchies among data. 
In this paper, we propose a \textbf{H}ier\textbf{a}rchical \textbf{b}i-directional self-attention \textbf{Net}work framework (HabNet) for paper review rating prediction and recommendation, which can serve as an effective decision-making tool for the academic paper review process.
Specifically, we leverage the hierarchical structure of the paper reviews with three levels of encoders: sentence encoder (level one), intra-review encoder (level two) and inter-review encoder (level three). 
Each encoder first derives contextual representation of each level, then generates a higher-level representation, and after the learning process, we are able to identify useful predictors to make the final acceptance decision, as well as to help discover the inconsistency between numerical review ratings and text sentiment conveyed by reviewers. 
Furthermore, we introduce two new metrics to evaluate models in data imbalance situations. 
Extensive experiments on a publicly available dataset (PeerRead) and our own collected dataset (OpenReview) demonstrate the superiority of the proposed approach compared with state-of-the-art methods.
\end{abstract}

\section{Introduction}\label{intro}
With an increasing submission of academic papers in recent years, the task of making final decisions manually incurs significant overheads to the program chairs, it is desirable to automate the process. 
In this study, we aim at utilizing document-level semantic analysis for paper review rating prediction and recommendation. 
Given the reviews of each paper from several reviewers as input, our goal is to infer the final acceptance decision for that paper and the reviewers' evaluation with respect to a numeric rating (e.g., 1-10 points). 
Paper review rating prediction and recommendation is a practical and important task in AI applications which will help improve the efficiency of the paper review process. It is also intended to enhance the consistency of the assessment procedures and outcomes, and to diversify the paper review process by comparing human recommended rating with machine recommended rating. 
In the literature, most of existing studies cast review rating prediction as a multi-class classification/regression task ~\cite{pang2005seeing}. 
They build a predictor by using supervised machine learning models with review texts and corresponding ratings. 
Due to the importance of features, most researches focus on extracting effective features such as context-level features \cite{qu2010bag} and user features \cite{gao2013modeling} to boost prediction performance. 
However, feature engineering is time-consuming and labor-intensive.

Recently, with the development of neural networks and its wide applications, various deep learning-based models have been proposed for automatically learning features from text data \cite{bengio2013representation}. 
Existing deep learning models usually learn continuous representations of different grains (e.g., word, phrase, sentence, document) from text corpus \cite{pennington2014glove,lai2015recurrent,kim2014convolutional,conneau2016very,wang2018disconnected,qiao2018new}. 
Although deep learning models can automatically learn extensive feature representation, they cannot efficiently capture the hierarchical relationship inherent to the review data. 
To address this problem, \newcite{yang2016hierarchical} studied a hierarchical architecture and implemented it in deep learning framework to learn a better document-level representation. 
Also, with the success of attention mechanism in many tasks such as machine translation, question answering and so on \cite{vaswani2017attention}, 
\newcite{shen2018disan} designed a directional self-attention network to gain context-aware embeddings for words and sentences. 
Despite great progress made by these models, they do not focus on the task of paper review rating recommendation and are not effective enough to be directly used for this task because of the following reasons:
First, the review data is hierarchical in nature. 
There exists a three-level hierarchical structure in the review data: word level, intra-review level and inter-review level, while previous models only capture two-levels (i.e., the word level and intra-review level) of this hierarchy. 
Second, paper reviews are usually much longer than other reviews (e.g., product reviews, movie reviews, restaurant reviews, etc.), while most of these models are working on those shorter reviews stated above and they do not leverage the up to date representation techniques such as BERT \cite{devlin2018bert} and SciBERT \cite{beltagy2019scibert}.

In this paper, we propose a novel neural network framework for paper review rating recommendation by taking word, intra-review and inter-review information into account. 
Specifically, inspired by HAN \cite{yang2016hierarchical} and DiSAN \cite{shen2018disan}, we introduce a Hierarchical Bi-directional self-Attention Network ({\methodname}) framework to effectively incorporate different levels of hierarchical information. 
The proposed framework consists of three main modules in end-to-end relationship: sentence encoder, intra-review encoder and inter-review encoder, which can consider hierarchical structures of review data as comprehensive as possible.
The outputs of inter-review encoder are leveraged as features to build the rating predictor without any feature engineering.
We release the code and data collected by us to enable replication and application to new tasks, available at \emph{https://github.com/RingBDStack/HabNet}.

The contributions of this work are as follows:
\begin{itemize}
\item We present a novel framework to guide the investigation and assessment of the effects of hierarchies on review data. 
To our best knowledge, this is the first work that incorporates different levels of semantic information into a hierarchical neural network to perform paper review rating recommendation.
\item We introduce two new metrics to better evaluate models when the distributions of classes are highly imbalanced (such as the paper review data we are working with).
\item Empirical results on OpenReview (ours) and extended PeerRead datasets demonstrates the effectiveness of the proposed method in automatically making final acceptance decisions and helping reveal the rating inconsistency between the semantic review content and the numerical review ratings.
\end{itemize}

\section{Related Work}
\subsection{Review Rating Prediction}
Review rating prediction is a basic task in sentiment analysis. 
It was initially studied by \newcite{pang2005seeing} who cast this problem as a multi-class classification/regression task. 
In the literature, most of studies following this approach used supervised machine learning models to do review rating prediction. 
Since the features used by these models are critical for prediction performance, more refined textual features are exploited. 
\newcite{qu2010bag} introduced bag of opinions representation, where an opinion was composed of a root word, a set of modifier words and one or more negation words.
\newcite{gao2013modeling} used user-specific and product-specific features to increase the reliability of sentiment classification. 
With the popularity of deep learning model, instead of hand-crafted features, many works were proposed to automatically learn features from text corpora.
\newcite{lai2015recurrent} applied a recurrent structure for convolutional neural network to capture contextual information for learning word representation. 
\newcite{conneau2016very} used very deep convolutional networks to learn hierarchical representations of whole sentences. 
\newcite{johnson2017deep} studied deepening word-level CNNs to capture global representations of text. 
\newcite{peng2018large} designed a deep Graph-CNN to learn both non-consecutive and long-distance features of text. 

\newcite{kangetal2018dataset} collected a dataset of peer reviews from several conferences and predicted paper acceptance decision by using paper draft.
\newcite{gaoetal2019rebuttal} focused on  predicting after-rebuttal scores by using their presented corpus. 
\newcite{huaetal2019argumentmining} applied argument mining on their AMPERE dataset to assess the efficiency of reviewing process. 
\newcite{li2019neural} designed a neural model to predict citation count of accepted papers.
\newcite{yang2018automatic} designed a hierarchical attention-based CNN for automatic academic paper rating by using source paper, it adopts original attention mechanism which cannot capture the interactions between elements in the same level. 
\newcite{leng2019deepreviewer} proposed DeepReviewer for automatic paper review utilizing paper's grammar and innovation to help learn better representation and predict paper's final review score. 
Different from above works, we aim at predicting the final acceptance decisions for papers and ratings for reviews with self-attention based framework using raw review texts. And our collected dataset contains the rating score of each review and the final decision of each paper.

\subsection{Attention Mechanism}
Attention mechanism was proposed by researchers to improve the performance of different NLP tasks.
There are two common attention mechanisms: additive attention \cite{bahdanau2014neural} and multiplicative attention \cite{rush2015neural,vaswani2017attention,peng2019hierarchical}, they use different compatibility functions to compute the attention weights. 
\newcite{lin2017structured} introduced self-attention to extract an interpretable sentence embedding. 
\newcite{yang2016hierarchical} proposed a hierarchical attention network for document classification, which applied attention mechanism at word and sentence level. 
\newcite {vaswani2017attention} built a simple network architecture based only on attention mechanism without convolutions and recurrence.
\newcite{yin2018attentive} proposed an attentive convolution network which enables deriving higher-level features for a word from information extracted from nonlocal context. 
\newcite{shen2018disan} designed a new attention mechanism which is directional and multi-dimensional, and a neural network solely based on this attention mechanism was proposed to learn sentence embedding. 
\newcite{shen2018bi} proposed a memory-efficient bi-directional self-attention network which splits sequence into blocks to save memory.

Our framework is also based on self-attention mechanism, which makes use of the hierarchical characteristic of HAN \cite{yang2016hierarchical} and the ability of capturing relationships between words from two directions in DiSAN \cite{shen2018disan}.

\section{Methodology}
In this section, we first describe the problem setting, and then present the details of our proposed framework for paper review rating prediction and recommendation.

\subsection{Problem Setting}
We consider the problem of paper review rating prediction and recommendation from a dataset containing $K$ papers, where each paper has $M$ reviews associated with the corresponding ratings and a decision class. Concretely, given the set $R = \{(r_1,c_1),...,(r_M,c_M), y\}$ for a scientific paper, where $r_i$ is the $i$-th reviewer's text review and $c_i$ is its associated numeric rating and $y$ is the final decision (i.e., accept or reject). Assume that each text review $r_i$ has $N$ sentences $S = \{\mathbf{s}_{i,1}, \mathbf{s}_{i,2}, \cdots, \mathbf{s}_{i,N}\}$ and each sentence contains $L$ words,
let $\mathbf{w}_{i, j, t}$ with  $i \in [1, M], j \in [1, N], t \in [1, L]$ denotes the $t$-th word in the $j$-th sentence of the $i$-th review document. Given a new paper with a set of reviews $R = \{r_1,...,r_M\}$, our goal is to predict the decision class $y$ which enables the program chairs to automatically make the final decision/recommendation, and also generate a rating $c$ for each review $r$ that is consistent with text sentiment as an aid to reviewers for discovering the rating inconsistency between ratings and review sentiments in the review process. 
Similar to \cite{zhang2010comparison,hassan2020multi}, here we treat paper review rating prediction problem as a multi-class classification problem, where the class labels are the rating scores $c$. We treat the final decision prediction as a binary classification problem, where the class labels are the decisions $y$.

\subsection{Our Approach}
\begin{figure}
    \centering
    \includegraphics[scale=0.76]{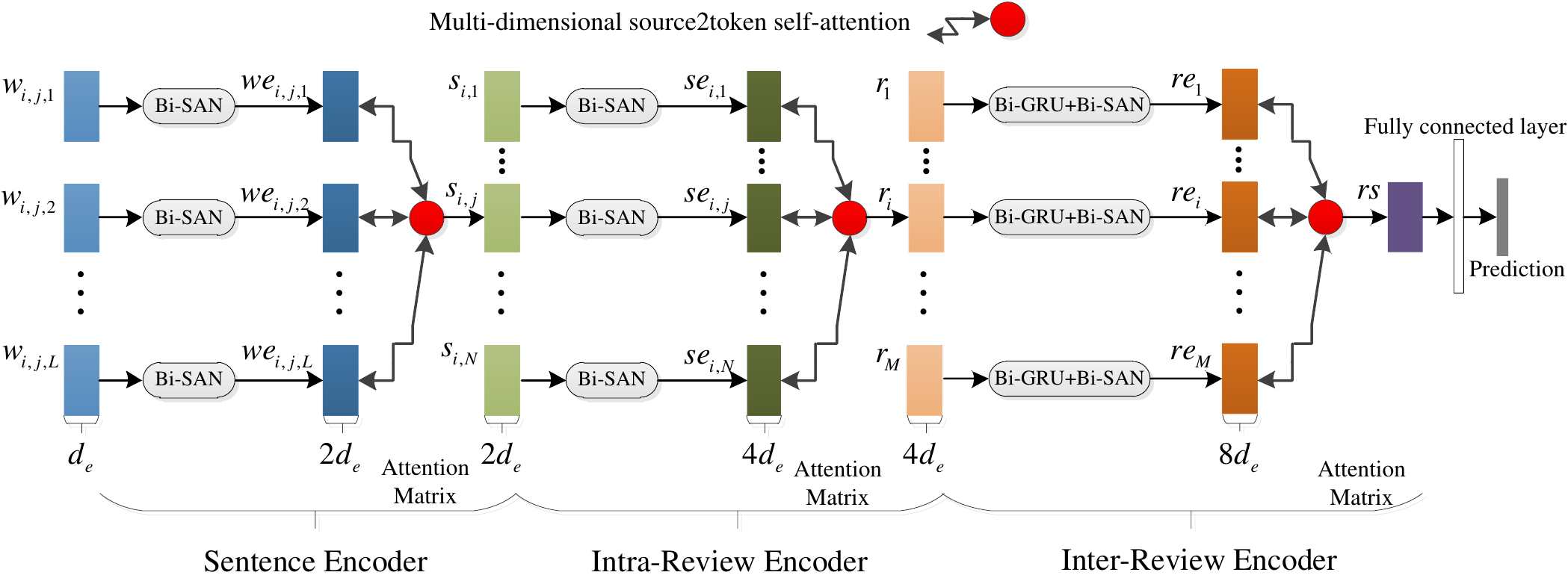}
    \caption{The architecture of HabNet framework.}
    \label{fig:framework_architecture}
\end{figure}

The proposed framework takes raw review texts as input and mainly consists of four components: sentence encoder, intra-review encoder, inter-review encoder and rating predictor, as shown in Figure \ref{fig:framework_architecture}.
Before describing the details of each component, we introduce the multi-dimensional source2token self-attention module by following \cite{shen2018disan} and taking this module in the sentence encoder as an example. The attention weight of each word $\mathbf {we}_{i,j,t}, t \in [1, L]$ is obtained by applying softmax on the scores $f(\mathbf{we}_{i,j,t}), t \in [1, L]$ calculated by Eq. (\ref{eq:source2token}), $W^T, W^{(1)}, b^{(1)}, b$ are trainable parameters. The output of this module is the weighted sum of the inputs (e.g., $\mathbf {we}_{i,j,t}, t \in [1, L]$ in sentence encoder).
\begin{equation}\label{eq:source2token}
    f(\mathbf{we}_{i,j,t}) = W^T\sigma(W^{(1)}\mathbf{we}_{i,j,t}+b^{(1)})+b.
\end{equation}

\noindent{\bf $\bullet$ \emph{Sentence Encoder.}}
Sentence encoder is designed to capture the relationships between words in a sentence and the importance of each word to the meaning of that sentence. It is shown in the first part of Figure \ref{fig:framework_architecture}.
It first generates context-aware embedding for each word in a sentence by using bi-directional self-attention module (Bi-SAN) \cite{shen2018disan}. 
Based on these context-aware embeddings of words, the encoding for that sentence, which contains all words' information and relations between words, is then obtained from the multi-dimensional source2token self-attention module \cite{shen2018disan} which aims at generating the sentence encoding by combining the context-aware word embeddings. 
Specifically, the input of sentence encoder are pre-trained word embeddings obtained from raw review texts by using GloVe pre-trained word embedding \cite{pennington2014glove}, or using BERT \cite{devlin2018bert} or SciBERT \cite{beltagy2019scibert}. 
Each word (e.g., $\mathbf{w}_{i,j,1}$, $\mathbf{w}_{i,j,2}$) is represented by a $d_e$-dimensional vector. 
These vectors are fed into Bi-SAN, which includes a forward self-attention network and a backward self-attention network. Each of these two networks outputs a refined embedding for each word and then the two refined embedding of each word are concatenated by Bi-SAN as the final context-aware embedding for each word (e.g., $\mathbf{we}_{i,j,1} \in \mathbb{R}^{2d_e}$).
The context-aware embedding for each word has $2d_e$ dimension because of the two networks (i.e., forward and backward) in Bi-SAN. 
After obtaining the context-aware embedding of each word, sentence encoder can generate encoding $\mathbf{s}_{i,j} \in \mathbb{R}^{2d_e}$ for each sentence through the multi-dimensional source2token self-attention module.

\noindent{\bf $\bullet$ \emph{Intra-Review Encoder.}}
Sentences in one review may have temporal orders, causality and other logic relationships, and some sentences contain more information for the review.
Therefore, intra-review encoder is designed to capture these relations existing in each individual review itself. 
The input is the sentence embedding $\mathbf{s}_{i,j}$ generated by the first-level sentence encoder. 
The structure of intra-review encoder is similar to sentence encoder where it first feeds sentence embedding to the Bi-SAN module, which captures the relations between sentences and the importance of one sentence to another from two directions by generating forward embedding $\mathbf{s}_{i,j}^{fw}$ and backward embedding $\mathbf{s}_{i,j}^{bw}$ for sentence $\mathbf{s}_{i,j}$. 
The final embedding $\mathbf{se}_{i,j} \in \mathbb{R}^{4d_e}$ for each sentence $\mathbf{s}_{i,j}$ in $i$-th review is generated by concatenating $\mathbf{s}_{i,j}^{fw}$ and $\mathbf{s}_{i,j}^{bw}$.
We have $\mathbf{se}_{i,j} =[\mathbf{s}_{i,j}^{fw}||\mathbf{s}_{i,j}^{bw}]$, where $||$ denotes concatenation operation.
Next, the multi-dimensional source2token self-attention module takes $\mathbf{se}_{i,j}$ as input and generates encoding $\mathbf{r}_i \in \mathbb{R}^{4d_e}$ for $i$-th review by combining all $\mathbf{se}_{i,j}$ in this review according to their importance weights, i.e., attention weights. As shown in the second part of Figure \ref{fig:framework_architecture}, intra-review encoder can generate encoding $\mathbf{r}_i$ for each review of the same paper. 
The dimension of $\mathbf{r}_i$ is $4d_e$, which is double of sentence encoding because of Bi-SAN.

\noindent{\bf $\bullet$ \emph{Inter-Review Encoder.}}
The integration of different reviews is essential for performing comprehensive analysis and supporting final decision-making on a paper. We use the inter-review encoder as the third level of our framework to integrate information from different reviews of each paper, as shown in the third part of Figure \ref{fig:framework_architecture}. It first feeds the second-level encoding $\mathbf{r}_i$ of $i$-th review of a paper to a bi-directional GRU \cite{bahdanau2014neural} layer, and then uses a Bi-SAN to model the relations between reviews from two directions by generating refined encoding  $\mathbf{re}_i$ for this review. 
Thus $\mathbf{re}_i$ contains the information from other reviews. Then, a multi-dimensional source2token self-attention module is applied on these encoding $\mathbf{re}_i$ to get a final compact vector representation $\mathbf{rs}$ of the paper. 
This encoder can handle papers having different number of reviews by using padding.
The whole process above is formulated as follows:

\textbf{Step1}: Feeding encoding $\mathbf{r}_i$ of each review of a paper generated by intra-review encoder to the bi-directional GRU layer, it outputs a new encoding for each review (we still use $\mathbf{r}_i$ to denote the new encoding of $i$-th review). Then these new encodings are fed to the following Bi-SAN module.

\textbf{Step2}: Bi-SAN has a forward self-attention network and a backward self-attention network. 
Two attention matrices, denoted as $\mathbf{P}^{i(fw)} \in \mathbb{R}^{4d_e\times M}$ and $\mathbf{P}^{i(bw)} \in \mathbb{R}^{4d_e\times M}$, for $i$-th review are calculated in these two networks respectively. 
Then the forward encoding $\mathbf{re}_i^{fw}$ and backward encoding $\mathbf{re}_i^{bw}$ for this review are generated as follows ($\odot$ denotes element-wise multiplication):
\begin{equation}\label{eq:forward&backward}
    \mathbf{re}_i^{fw} = \sum_{o=1}^M \mathbf{P}^{i(fw)}_{\cdot o}\odot \mathbf{r}_o, ~~ \mathbf{re}_i^{bw} = \sum_{o=1}^M \mathbf{P}^{i(bw)}_{\cdot o}\odot \mathbf{r}_o,
\end{equation}
where $M$ is the number of reviews for one paper.
$\mathbf{P}^{i(fw)}_{\cdot o}$ and $\mathbf{P}^{i(bw)}_{\cdot o}$ denote the $o$-th column in attention matrix $\mathbf{P}^{i(fw)}$ and $\mathbf{P}^{i(bw)}$ respectively. 
The refined encoding $\mathbf{re}_i$ for $i$-th review, which contains the information from other reviews of the same paper, is generated in the following equation.
\begin{equation}\label{eq:re}
    \mathbf{re}_i =[\mathbf{re}_i^{fw}|| \mathbf{re}_i^{bw}], \mathbf{re}_i \in \mathbb{R}^{8d_e}.
\end{equation}

\textbf{Step3}: The multi-dimensional source2token self-attention module takes the encodings of all reviews of one paper outputed from Bi-SAN as input, and computes the importance weight for each review encoding $\mathbf{re}_i$, and then combines all these review encodings to get the final vector representation $\mathbf{rs}$ of the paper based on the importance weights in the similar way as shown in Eq. (\ref{eq:forward&backward}).

\noindent{\bf $\bullet$ \emph{Rating Prediction and Recommendation.}}
With the three levels of encoding above, a fully connected layer with softmax function is designed to make rating prediction and final recommendation. Specifically, we take the compact representation $\mathbf{rs}$ from all reviews as its input to predict the final decision, and the encoding $\mathbf{r}_i$ for $i$-th review as its input to predict the corresponding rating, respectively. It is worth noting that the predicted review ratings are consistent with text sentiment conveyed by reviewers, thus it can serve as a guidance to reviewers for finding the inconsistencies between semantic review content and numerical review ratings in the review process.

\subsection{Model Variants}
To understand the contribution of different components in the proposed framework, we derive different variants for ablation study. Below are three variants implemented in our experiments.

\textbf{{\methodname}-V1:} After obtaining the encoding $\mathbf{r}_i$ of each review for a paper which is outputed from intra-review encoder, we sum them up using equal weight and then use the result as the final encoding $\mathbf{rs}$ of that paper, i.e., $\mathbf{rs} = \frac{1}{M}\sum_{i=1}^M\mathbf{r}_i$. 
Thus the inter-review encoder is removed in this variant.

\textbf{{\methodname}-V2:} We remove the sentence encoder in the proposed framework as the second variant to verify the contribution of sentence encoder to the framework.
Specifically, for a sentence, we use the average of all words' pre-trained embeddings as its encoding, and feed such sentence encodings to intra-review encoder. 
Therefore, this variant cannot encode the relations between words in a sentence.

\textbf{{\methodname}-V3:} We remove the intra-review encoder as the third variant to understand how effectively this encoder captures the interactions between sentences in a review document and to demonstrate the importance of intra-review encoder to the proposed framework. 
To be specific, the encoding of a review document is the mean of sentence embeddings in that review.

\section{Experiments and Results}
\subsection{Dataset}
We conduct experiments on two datasets to validate our approach for scientific paper decision recommendation and review rating prediction.
One is called OpenReview dataset which is collected by us. 
The other one is a dataset extended from PeerRead which is originally published by \newcite{kangetal2018dataset}. Table \ref{table:stats_of_datasets} shows the statistics of OpenReview and Extended PeerRead.
For all the experiments on these two datasets, all samples are randomly shuffled before splitting the dataset.

\noindent{\bf $\bullet$ \emph{OpenReview.}}
This collection contains all reviews for ICLR conference and workshop's papers from 2017 to 2019.
Generally, each paper has 3-5 reviews with corresponding ratings, the rating is a numeric value from 1 to 10 (10 being the highest rating). There is also a decision (accept or reject) associated with each paper.
The number of accepted and rejected papers are 1341 and 1962. 
For paper decision recommendation, we use 2293, 491 and 492 papers as training, validation and testing set respectively.
For review rating prediction, 7600, 1000, 1000 reviews are used as training, validation and testing set respectively.
As shown in Table \ref{table:stats_of_datasets}, the number of reviews with different ratings are highly imbalanced.

\noindent{\bf $\bullet$ \emph{Extended PeerRead.}}
The majority of papers with reviews in the original PeerRead dataset are accepted papers collected from NIPS 2013-2017, as shown in Table \ref{table:stats_of_datasets}, the number of accepted and rejected papers are 2054 and 0, respectively. Thus the original PeerRead dataset cannot be used directly for predicting final decisions on accepted/unaccepted papers due to the severe imbalance problem stated above. Therefore, we further collect 2211 papers from ICLR 2020 conference and corresponding reviews from the openreview website to extend the PeerRead dataset.
Finally, the extended PeerRead dataset has 4265 papers and 13721 reviews in total. However, since most review ratings are not available in the original PeerRead, we only use this extended dataset to predict the final decision. 

\newcommand{\tabincell}[2]{\begin{tabular}{@{}#1@{}}#2\end{tabular}} 

\begin{table}[h!]
\small
\centering

\begin{tabular}{m{1.4cm}|m{2.2cm} m{1.0cm} m{1.1cm} m{1.3cm}|m{0.2cm} m{0.2cm} m{0.2cm} m{0.5cm} m{0.5cm} m{0.5cm} m{0.5cm} m{0.2cm} m{0.2cm} m{0.2cm}} 
 \hline
 \multirow{2}{*}{Dataset} & \multirow{2}{*}{Section} & \multirow{2}{*}{\#Papers} & \multirow{2}{*}{\#Reviews} & \multirow{2}{*}{\#Acc/Rej} & \multicolumn{10}{c}{Review Rating Distribution} \\
 \cline{6-15}
 &  & & & & 1 & 2 &3 & 4 & 5 & 6 & 7 & 8 & 9 & 10 \\
 \hline
   OpenReview & ICLR 2017-2019 & 3,303 & 9,600 & 1,341/1,962 & 37 & 205 & 851 & 1,816 & 1,943 & 2,154 & 1,875& 563 & 158 & 16 \\ [0.5ex]
 \hline
 \multirow{3}{0.2cm}{Extended PeerRead} & NIPS 2013-2017 & 2,054 & 7,006 & 2,054/0 & - \\ [0.5ex]
 & ICLR 2020 & 2,211 & 6,715 & 687/1,524 & - \\ [0.5ex]
 \cline{2-15}
 & Total & 4,265 & 13,721 & 2,741/1,524 & - \\ [0.5ex]
\hline
\end{tabular}
\caption{Statistics of OpenReview and Extended PeerRead, - means the rating distribution is unavailable.}
\label{table:stats_of_datasets}
\end{table}

\subsection{Evaluation Metrics and Baselines}
We use Accuracy, Macro-F1 and Micro-F1 to evaluate the effectiveness of our framework on the task of paper decision recommendation. 
For review rating prediction, due to the imbalanced distribution of ratings (shown in Table \ref{table:stats_of_datasets}) and ineffectiveness of methods dealing with imbalanced problem (such as the oversampling technique and reducing rating range we tried), two new metrics with better discernibility are designed to better evaluate the performance of our framework and baselines apart from Accuracy.

\textbf{Distance Measure (DM).} The distance between true label and predicted label is crucial for evaluating a model when there are multiple labels as in our task of review rating prediction. 
The smaller the distance, the better a model works. 
Thus we design a new metric which incorporates the distance between predicted rating and true rating. 
This metric can distinguish a better model from a more reasonable perspective. 
For example, models which predict a rating of 8 as 7 are much better than models that predict it as 3. 
Let $p_i$ and $r_i$ be the predicted rating and true rating for $i$-th sample respectively, and $n$ be the total number of samples. 
We define $DM$ as follows: 
\begin{equation}\label{eq:dm}
    DM = \frac{1}{n}\sum_{i=1}^n(1-\frac{d_i}{d_{max}}),\quad \text{where}\quad d_i = |p_i - r_i|.
\end{equation}

It first calculates the distance $d_i$ for each sample and then takes an average over all samples according to Eq. (\ref{eq:dm}). 
When the predictions for all samples are correct, the value of $DM$ achieves its best which is $1$. 
When all predictions are wrong and the distances between predicted ratings and true ratings are all maximum distances $d_{max}$ (in our case $d_{max}=9$), the value of $DM$ is $0$. 
When the distances become smaller, the value of $DM$ becomes larger. 
Thus it can evaluate the performance of models appropriately. 
The range of $DM$'s value is $[0,1]$. The larger its value is, the better the algorithm works.

\textbf{Optimized Precision (OP).} It is important to correctly predict all classes when the data is imbalanced. Inspired by \newcite{hossin2015review}, we combine accuracy and recall of all classes into a unified measure, which allows to better deal with imbalanced data environments.
Let $ACC$ be the accuracy,
$N$ be the number of classes, and $R_i$ be the recall for $i$-th class, $i=1, \cdots, N$. We define $OP$ as follows:
\begin{equation}\label{eq:op}
    OP = ACC -\frac{\sum_{i,j=1}^N|R_i-R_j|}{2(N-1)\sum_{k=1}^N R_k}.
\end{equation}
As shown in Eq. (\ref{eq:op}), $OP$ first computes the absolute differences between recalls of each pair of classes and sum them up, and then normalizes it by using the sum of all recalls. 
In this way, this metric measures the model's ability to predict the highest score of both accuracy and recall for all classes. The higher the value of $OP$, the better the model fits the data.

We compare our proposed method with fourteen other state-of-the-art methods and three variations of our proposed model: (1). 10 flat baselines: three are traditional text classification models, including Support Vector Machines (SVM), Logistic Regression (LR) and Naïve Bayes (NB); five are deep learning models, including RNN (i.e., Bi-GRU) \cite{cho2014learning,bahdanau2014neural}, TextCNN \cite{kim2014convolutional}, TextRCNN \cite{lai2015recurrent}, VDCNN \cite{conneau2016very} and DPCNN \cite{johnson2017deep}; two are attention-based models, including Transformer \cite{vaswani2017attention} and SA-Sent-EM \cite{lin2017structured}, which exploit various relationships existing in review text.
(2). 4 hierarchical baselines which leverage the hierarchical structure of the dataset: HAN-extended is an extension of HAN \cite{yang2016hierarchical} re-implemented by us; we also implement three Bert-based baselines \cite{devlin2018bert,beltagy2019scibert} using large pre-trained contextual embeddings. Specifically, Bert-base and Bert-large use 768 and 1024-dimensional Bert embedding respectively, while SciBert utilizes 768-dimensional SciBert embedding. For our proposed framework {\methodname}, apart from using GloVe embedding, we also conduct experiments by using the above three Bert-based contextual embeddings. (3). To demonstrate the contribution of each encoder ingredient, we also implement three variants of our proposed framework.

\subsection{Experimental Settings}
We use raw review texts as input for all models.
For the decision recommendation task, 50-dimensional pre-trained GloVe word embedding is used for the models of our framework and HAN-extended, while 100-dimensional one is adopted for other deep learning models except bert-based baselines which utilize corresponding bert-embeddings.
The number of training epochs is set to 100.
For the rating prediction task, except bert-based baselines using bert embeddings, 100-dimensional pre-trained GloVe word embedding is used for all models. The number of training epochs is set to 50 since all models converge quickly.
We use cross entropy as objective function to train all deep learning models. 
The common parameters, such as learning rate and batch size, are empirically set.
For all the experiments on {\methodname} and its variants, we train each of them 10 times and use the average results to evaluate them.

\subsection{Experimental Results and Discussion}
The experimental results are shown in Table \ref{table:results_binary_cls}. 
For the paper decision recommendation task, {\methodname} achieves the best performance results
no matter which kind of embedding is used. 
This demonstrates the effectiveness of our framework and its generality. 
To be specific,
compared with flat baselines, our framework with GloVe embedding, i.e., {\methodname} (Glove), performs much better, which demonstrates that our framework can make good use of the hierarchical structure in the dataset.
While compared with hierarchical baselines, {\methodname} (Bert-base), {\methodname} (Bert-large) and {\methodname} (SciBert) obtain good performance gain (5.4\%, 8.2\%, 5.4\% and 8.2\%, 8.0\%, 4.1\% in terms of accuracy on both datasets) over corresponding bert-based baseline respectively.
The improvement indicates that our framework can capture the relationships between words, sentences, and reviews existing in the dataset. Although the three bert-based baselines can obtain contextual word embeddings, they cannot capture intra-review level and inter-review level relationships as our framework does. 
Even {\methodname} (Glove) still performs better than the three bert-based baselines using pre-trained contextual embedding and HAN-extended, which further demonstrates the ability of the encoders on capturing the three-level relationships.
In addition, the best performance of our framework on both datasets validates its generality, and {\methodname} with different embedding outperforming all the baselines consolidates this.
It is worth noting that the performance results of all models on the extended PeerRead dataset are higher than that on the OpenReview dataset. The reason may be that the review texts (especially those from NIPS 2013-2017) in the PeerRead dataset are much shorter and less complex than those in the OpenReview dataset. This demonstrates that our framework can work on long review text much better than other models.

Note that in Table \ref{table:results_binary_cls}, there are only results on OpenReview dataset for review rating prediction, as PeerRead does not contain ratings.
{\methodname} with various embedding (including GloVe and bert embeddings) achieving the best performance demonstrates the effectiveness and generalization ability of our framework again, because {\methodname} has a similar performance improvement as in the paper decision recommendation task when compared with flat and hierarchical baselines.
Furthermore, the ratings predicted by {\methodname}, although not completely correct, can still be used as an aid to find inconsistencies between given ratings and text sentiments conveyed by reviewers.

\begin{table}
\small
\centering
\begin{tabular}{c|c| c c c| c c c| c c c} 
 \hline
		\multirow{3}{*}{}  & \multirow{3}{*}{Models}    &\multicolumn{6}{c|}{Scientific Paper Decision Recommendation} & \multicolumn{3}{c}{Review Rating Prediction} \\ \cline{3-11} 
		& & \multicolumn{3}{c|}{OpenReview} & \multicolumn{3}{c|}{Extended PeerRead} & \multicolumn{3}{c}{OpenReview} \\ \cline{3-11}
		& & ACC & Ma-F1 & Mi-F1 & ACC & Ma-F1 & Mi-F1 & ACC & DM & OP \\ [0.5ex] 
 \hline
 \multirow{10}{0.4cm}{\rotatebox{90}{\tabincell{c}{Flat Baselines}}}
 & NB & 0.599 & 0.375 & 0.449 & 0.643 & 0.391 & 0.504 & 0.225 & 0.854 & -0.772 \\ [0.5ex]
 & LR & 0.699 & 0.635 & 0.671 & 0.794 & 0.763 & 0.789 & 0.316 & 0.883 & -0.401 \\ [0.5ex]
 & SVM & 0.686 & 0.603 & 0.643 & 0.790 & 0.757 & 0.783 & 0.318 & 0.883 & -0.392 \\ [0.5ex]
 \cline{2-11}
 & RNN \cite{bahdanau2014neural} & 0.606 & 0.404 & 0.472 & 0.697 & 0.628 & 0.679 & 0.250 & 0.856 & -0.318 \\ [0.5ex]
 & TextCNN \cite{kim2014convolutional} & 0.677 & 0.606 & 0.638 & 0.820 & 0.802 & 0.821 & 0.284 & 0.861 & -0.356 \\ [0.5ex]
 & TextRCNN \cite{lai2015recurrent} & 0.648 & 0.595 & 0.624 & 0.816 & 0.803 & 0.819 & 0.271 & 0.854 & -0.364 \\ [0.5ex]
 & VDCNN \cite{conneau2016very} & 0.616 & 0.551 & 0.584 & 0.667 & 0.456 & 0.565 & 0.233 & 0.849 & -0.350 \\ [0.5ex]
 & DPCNN \cite{johnson2017deep} & 0.642 & 0.478 & 0.551 & 0.831 & 0.828 & 0.835 & 0.295 & 0.874 & -0.230 \\ [0.5ex]
  \cline{2-11}
 & Transformer \cite{vaswani2017attention} & 0.602 & 0.381 & 0.458 & 0.720 & 0.658 & 0.705 & 0.212 & 0.841 & -0.322 \\ [0.5ex]
 & SA-Sent-EM \cite{lin2017structured} & 0.699 & 0.662 & 0.683 & 0.831 & 0.821 & 0.834 & 0.323 & 0.885 & -0.204 \\ [0.5ex]
 \hline
 \multirow{4}{0.4cm}{\rotatebox{90}{\tabincell{c}{Hierarchical\\ Baselines}}}
 & HAN \cite{yang2016hierarchical}-extended & 0.713 & 0.709 & 0.716 & 0.833 & 0.816 & 0.834 & 0.338 & 0.887 & -0.187 \\ [0.5ex]
 \cline{2-11}
 & Bert-base \cite{devlin2018bert} & 0.735 & 0.702 & 0.721 & 0.814 & 0.806 & 0.817 & 0.331 & 0.887 & -0.208 \\ [0.5ex]
 & Bert-large \cite{devlin2018bert} & 0.736 & 0.706 & 0.725 & 0.816 & 0.803 & 0.816 & 0.330 & 0.884 & -0.270 \\ [0.5ex]
 & SciBert \cite{beltagy2019scibert} & 0.746 & 0.730 & 0.743 & 0.845 & 0.831 & 0.844 & 0.341 & 0.887 & -0.176 \\ [0.5ex]
 \hline
 \multirow{4}{0.4cm}{\rotatebox{90}{\tabincell{c}{Ours}}}
 & {\methodname} (Glove) & \textbf{0.753} & \textbf{0.730} & \textbf{0.745} & \textbf{0.876} & \textbf{0.863} & \textbf{0.877} & \textbf{0.356} & \textbf{0.890} & \textbf{-0.061} \\ [0.5ex]
 & {\methodname} (Bert-base) & \textbf{0.775} & \textbf{0.766} & \textbf{0.776} & \textbf{0.881} & \textbf{0.870} & \textbf{0.880} & \textbf{0.375} & \textbf{0.901} & \textbf{0.019} \\ [0.5ex]
 & {\methodname} (SciBert) & \textbf{0.786} & \textbf{0.779} & \textbf{0.787} & \textbf{0.880} & \textbf{0.869} & \textbf{0.879} & \textbf{0.365} & \textbf{0.899} & \textbf{-0.013} \\ [0.5ex]
 & {\methodname} (Bert-large) & \textbf{0.796} & \textbf{0.787} & \textbf{0.796} & \textbf{0.881} & \textbf{0.873} & \textbf{0.882} & \textbf{0.379} & \textbf{0.900} & \textbf{0.032} \\ [0.5ex]
\hline
\end{tabular}
\caption{Performance results of all models on OpenReview and Extended PeerRead datasets. ``ACC", ``Ma-F1" and ``Mi-F1" denote Accuracy, Macro-F1 and Micro-F1 respectively.}
\label{table:results_binary_cls}
\end{table}

\subsection{Ablation Study}
We conduct ablation study of our framework to evaluate the contribution of each component. 
The results are shown in Table \ref{table:ablation_study}.
For the paper decision recommendation task, the better performance of {\methodname} over {\methodname}-V1 on both datasets indicates that the inter-review encoder can integrate information from different reviews of one paper well which helps the decision recommendation. While {\methodname} performs better than {\methodname}-V2 verifies the importance of sentence encoder which can encode the relationships between words in a sentence. And {\methodname} outperforming {\methodname}-V3 demonstrates the ability of intra-review encoder to capture sentence-level relations in a review text and that such relations between sentences contribute much information to the meaning of the review document. The results of {\methodname} and the variants on the review rating prediction task have a similar trend which further validates the contribution of different encoders to the framework. In conclusion, the three encoders help {\methodname} capture three-level relationships in the dataset which plays a vital role on improving prediction performance.

\begin{table}[h!]
\small
\centering

\begin{tabular}{m{0.5cm}|m{1.5cm}|m{1.3cm} m{1.3cm} m{1.3cm}|m{1.3cm} m{1.3cm}|m{1.3cm} m{1.3cm} m{1.3cm}}
 \hline
    \multirow{2}{*}{Task}  & \multirow{2}{*}{Models} & \multicolumn{5}{c|}{OpenReview Dataset} & \multicolumn{3}{c}{Extended PeerRead Dataset} \\ \cline{3-10}
 &  & Accuracy & Macro-F1 & Micro-F1 & DM & OP & Accuracy & Macro-F1 & Micro-F1 \\
 \hline
 \multirow{4}{0.2cm}{\rotatebox{90}{\tabincell{c}{Decision\\ Prediction}}} & {\methodname}-V1 & 0.735 & 0.705 & 0.723 & - & - & 0.858 & 0.846 & 0.860 \\ [0.5ex]
   & {\methodname}-V2 & 0.736 & 0.716 & 0.730 & - & - & 0.861 & 0.843 & 0.860 \\ [0.5ex]
   & {\methodname}-V3 & 0.726 & 0.700 & 0.717 & - & - & 0.859 & 0.846 & 0.861 \\ [0.5ex]
   & {\methodname} & \textbf{0.753} ($\uparrow$) & \textbf{0.730} ($\uparrow$) & \textbf{0.745} ($\uparrow$) & - & - & \textbf{0.876} ($\uparrow$) & \textbf{0.863} ($\uparrow$) & \textbf{0.877} ($\uparrow$) \\ [0.5ex]
 \hline
 \multirow{3}{0.2cm}{\rotatebox{90}{\tabincell{c}{Rating\\ Prediction}}} & {\methodname}-V2 & 0.335 & - & - & 0.886 & -0.218 & - & - & - \\ [0.5ex]
 & {\methodname}-V3 & 0.336 & - & - & 0.887 & -0.210 & - & - & - \\ [0.5ex]
 & {\methodname} & \textbf{0.356} ($\uparrow$) & - & - & \textbf{0.890} ($\uparrow$) & \textbf{-0.061} ($\uparrow$) & - & - & - \\ [0.5ex]
\hline
\end{tabular}
\caption{Results of ablation study of our framework on OpenReview and Extended PeerRead datasets. Accuracy, Macro-F1 and Micro-F1 are the metrics used for the decision prediction/recommendation task; while Accuracy, DM and OP are used for the rating prediction task, and there are no results for this task on the extended PeerRead dataset because this dataset does not have rating for each review. {\methodname} achieves the best results which are in bold, arrow $\uparrow$ indicates statistical significance ($p < 0.05$).}
\label{table:ablation_study}
\end{table}

\subsection{Case Study}
\begin{figure}[t]
    \centering
    \includegraphics[height = 1.3in,width=5.15in]{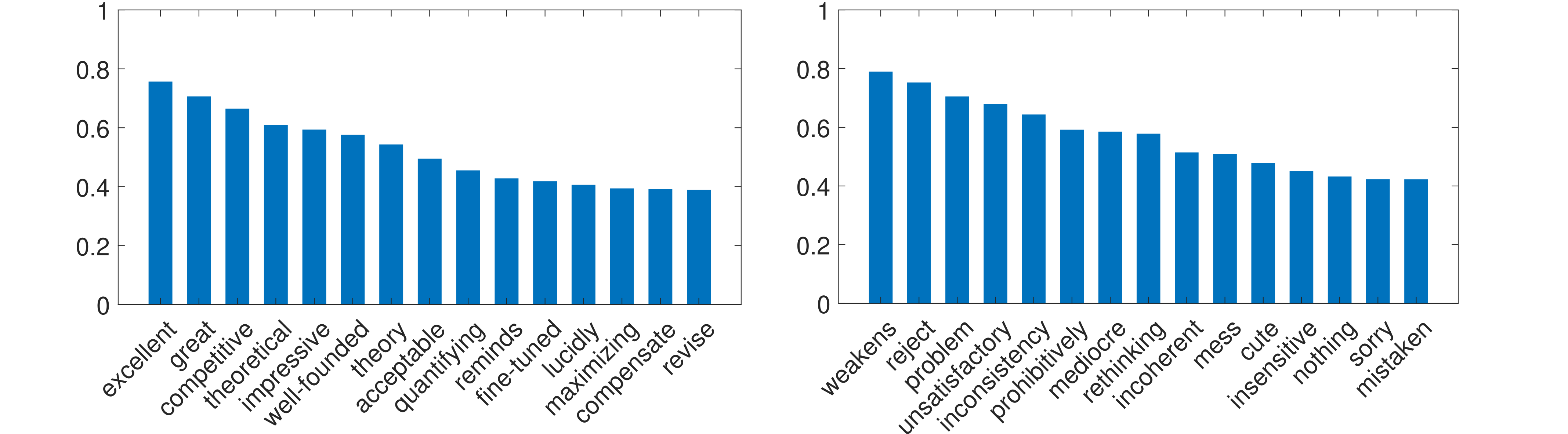}
    \caption{Attention weights of the top 15 approbatory words for accepted and rejected papers, left sub-figure is for accepted papers, right one is for rejected papers.}
    \label{fig:top15words}
\end{figure}

To gain a view over the ability of our proposed framework on capturing the importance of words in the scientific paper decision recommendation task, we visualize the top-15 approbatory words for accepted and rejected papers, as shown in Figure~\ref{fig:top15words}.
One can see that for the accepted papers, the attention on positive words such as ``excellent", ``competitive" and so on are much higher than other words.
For the rejected papers, negatives words such as ``unsatisfactory", ``incoherent" and so on have higher attention weight than others. 
Intuitively, reviewers can express their tendency towards the result of article more clearly through the above keywords.
Moreover, compared to the attention weights of words in accepted papers, the attention weights of words in rejected papers are generally greater.
The possible reason is that reviewers' comments on the rejected articles are more consistent than that of accepted articles.

\begin{figure*}[htbp]
\centering
\subfigure[A review for an accepted paper.]{
    \begin{minipage}[t]{1.0\linewidth}
        \centering
        \includegraphics[height = 0.95in,width=6in]{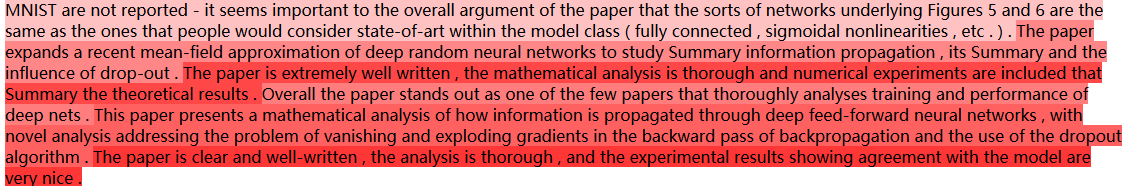}
    \end{minipage}%
}%

\subfigure[A review for a rejected paper.]{
    \begin{minipage}[t]{1.0\linewidth}
        \centering
        \includegraphics[height = 1.7in,width=6in]{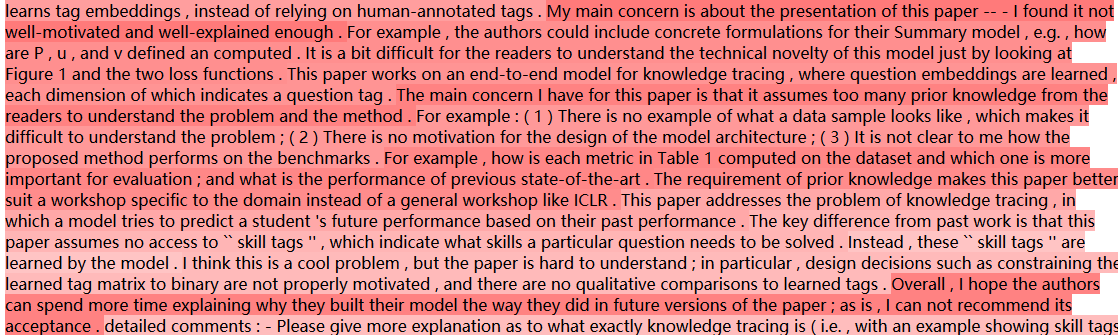}
    \end{minipage}%
}%
\centering
\caption{Sentence-level attention visualization for accepted and rejected papers.}
\label{fig:sent-level-attn}
\end{figure*}

We also visualize the sentence-level attention on one accepted paper and rejected paper respectively, as shown in Figure \ref{fig:sent-level-attn}, the deeper the color, the bigger the attention weight. For the accepted paper, the sentence with the deepest color expresses strong positive attitude towards the acceptance decision, while other sentences without strong sentiment have smaller attention weights (i.e., the color is much lighter). The same trend also appears in the rejected paper. This result shows that our framework can capture the most important sentence-level signal within a review for predicting the final decisions for papers.

\subsection{Error Analysis}
We investigate the error cases that {\methodname} did not predict correctly on OpenReview dataset and find that: (1) 67\% of them are predicted by {\methodname} as rejected papers, but they are actually accepted paper; (2) 33\% of them are predicted as accepted but are in fact rejected. 

We randomly select 20 examples from (1) and read the review text carefully, we find that: a). 18 of them have many negative keywords and phrases, like ``unclear", ``limited", ``hard to interpret", ``not provable" and so on. Although there are also positive expressions such as ``looks good", ``interesting", majority of the content contains negative ones. Thus the overall sentiment of the review text is classified as negative by {\methodname}.
b). 2 of them have indicator of acceptance, like the keyword ``accepted" or ``acceptance", but meanwhile there are also many negative words, such as ``confusing", ``no comparison". All of these positive and negative information together make the model unable to make correct prediction.

We also randomly select 20 examples from (2). There are three cases: a). 7 of them contain very strong acceptance keywords and sentences, such as ``pretty impressive", ``promising", ``I recommend acceptance". Because of these strong indicators, {\methodname} predict them as accepted papers. b). 2 of them have strong indicators of rejection, such as ``The novelty of the paper is not enough to justify its acceptance", but they also have several strong positive keywords which deviate the overall sentiment of the review text and thus affect {\methodname}'s prediction.
c). 11 of them have many positive and negative keywords and sentences at the same time, and there is no strong indicator of rejection. {\methodname} can not deal with them very well, because it takes all the positive and negative information into consideration.

\section{Conclusion}
In this paper, a scientific paper review dataset called OpenReview is collected from ICLR openreview website and released.
We observe that there is a three-level hierarchical structure in this dataset (i.e., word level, intra-review level and inter-review level) -- the information and relationships between reviews of one paper may affect the final decision, and so may relationships between words and sentences in each review.
Based on these observations, a hierarchical bi-directional self-attention network (HabNet) framework is proposed for paper review rating prediction and recommendation that can model the interactions among words, sentences, intra- and inter-reviews in an end-to-end manner. Moreover, considering the imbalanced distribution of different classes (i.e., ratings from 1 to 10) in the review rating prediction task, we design two new metrics to better evaluate models. 
It is seen that both experimental results of predicting final decisions for submitted papers and identifying ratings for reviews on two datasets (OpenReview and extended PeerRead) demonstrate our proposed framework has sufficient ability to capture the hierarchical structures of words, sentences and reviews in the datasets and outperforms other models.
In the future, we plan to investigate multi-task learning for paper review rating recommendation.

\section*{Acknowledgements}
The corresponding author is Hao Peng. This work is supported by the NSFC NO.62002007 and 61872022, NSF of Guangdong Province (2017A030313339), and in part by NSF under grants III-1763325, III-1909323, and SaTC-1930941. 
We thank the reviewers for their constructive comments.


\bibliographystyle{coling}
\bibliography{coling2020}

\begin{thebibliography}{}

\bibitem[\protect\citename{Bahdanau \bgroup et al.\egroup
  }2015]{bahdanau2014neural}
Dzmitry Bahdanau, Kyunghyun Cho, and Yoshua Bengio.
\newblock 2015.
\newblock Neural machine translation by jointly learning to align and
  translate.
\newblock In {\em Proceedings of the ICLR}.

\bibitem[\protect\citename{Beltagy \bgroup et al.\egroup
  }2019]{beltagy2019scibert}
Iz~Beltagy, Kyle Lo, and Arman Cohan.
\newblock 2019.
\newblock Scibert: A pretrained language model for scientific text.
\newblock In {\em Proceedings of the EMNLP}, pages 3606--3611.

\bibitem[\protect\citename{Bengio \bgroup et al.\egroup
  }2013]{bengio2013representation}
Yoshua Bengio, Aaron Courville, and Pascal Vincent.
\newblock 2013.
\newblock Representation learning: A review and new perspectives.
\newblock {\em TPAMI}, 35(8):1798--1828.

\bibitem[\protect\citename{Cho \bgroup et al.\egroup }2014]{cho2014learning}
Kyunghyun Cho, Bart van Merri{\"e}nboer, Caglar Gulcehre, Dzmitry Bahdanau,
  Fethi Bougares, Holger Schwenk, and Yoshua Bengio.
\newblock 2014.
\newblock Learning phrase representations using rnn encoder--decoder for
  statistical machine translation.
\newblock In {\em Proceedings of the EMNLP}, pages 1724--1734.

\bibitem[\protect\citename{Conneau \bgroup et al.\egroup
  }2017]{conneau2016very}
Alexis Conneau, Holger Schwenk, Lo{\"\i}c Barrault, and Yann Lecun.
\newblock 2017.
\newblock Very deep convolutional networks for text classification.
\newblock In {\em Proceedings of the EACL}, pages 1107--1116.

\bibitem[\protect\citename{Devlin \bgroup et al.\egroup }2019]{devlin2018bert}
Jacob Devlin, Ming-Wei Chang, Kenton Lee, and Kristina Toutanova.
\newblock 2019.
\newblock Bert: Pre-training of deep bidirectional transformers for language
  understanding.
\newblock In {\em Proceedings of the NAACL}, pages 4171--4186.

\bibitem[\protect\citename{Gao \bgroup et al.\egroup }2013]{gao2013modeling}
Wenliang Gao, Naoki Yoshinaga, Nobuhiro Kaji, and Masaru Kitsuregawa.
\newblock 2013.
\newblock Modeling user leniency and product popularity for sentiment
  classification.
\newblock In {\em Proceedings of the IJCNLP}, pages 1107--1111.

\bibitem[\protect\citename{Gao \bgroup et al.\egroup
  }2019]{gaoetal2019rebuttal}
Yang Gao, Steffen Eger, Ilia Kuznetsov, Iryna Gurevych, and Yusuke Miyao.
\newblock 2019.
\newblock Does my rebuttal matter? insights from a major nlp conference.
\newblock In {\em Proceedings of the NAACL}, pages 1274--1290.

\bibitem[\protect\citename{Hassan and Shoaib}2020]{hassan2020multi}
Junaid Hassan and Umar Shoaib.
\newblock 2020.
\newblock Multi-class review rating classification using deep recurrent neural
  network.
\newblock {\em NPL}, 51(1):1031--1048.

\bibitem[\protect\citename{Hossin and Sulaiman}2015]{hossin2015review}
Mohammad Hossin and MN~Sulaiman.
\newblock 2015.
\newblock A review on evaluation metrics for data classification evaluations.
\newblock {\em IJDKP}, 5(2):1.

\bibitem[\protect\citename{Hua \bgroup et al.\egroup
  }2019]{huaetal2019argumentmining}
Xinyu Hua, Mitko Nikolov, Nikhil Badugu, and Lu~Wang.
\newblock 2019.
\newblock Argument mining for understanding peer reviews.
\newblock In {\em Proceedings of the NAACL}, pages 2131--2137.

\bibitem[\protect\citename{Johnson and Zhang}2017]{johnson2017deep}
Rie Johnson and Tong Zhang.
\newblock 2017.
\newblock Deep pyramid convolutional neural networks for text categorization.
\newblock In {\em Proceedings of the ACL}, pages 562--570.

\bibitem[\protect\citename{Kang \bgroup et al.\egroup
  }2018]{kangetal2018dataset}
Dongyeop Kang, Waleed Ammar, Bhavana Dalvi, Madeleine van Zuylen, Sebastian
  Kohlmeier, Eduard Hovy, and Roy Schwartz.
\newblock 2018.
\newblock A dataset of peer reviews (peerread): Collection, insights and nlp
  applications.
\newblock In {\em Proceedings of the NAACL}, pages 1647--1661.

\bibitem[\protect\citename{Kim}2014]{kim2014convolutional}
Yoon Kim.
\newblock 2014.
\newblock Convolutional neural networks for sentence classification.
\newblock In {\em Proceedings of the EMNLP}, pages 1746--1751.

\bibitem[\protect\citename{Lai \bgroup et al.\egroup }2015]{lai2015recurrent}
Siwei Lai, Liheng Xu, Kang Liu, and Jun Zhao.
\newblock 2015.
\newblock Recurrent convolutional neural networks for text classification.
\newblock In {\em Proceedings of the AAAI}, pages 2267--2273.

\bibitem[\protect\citename{Leng \bgroup et al.\egroup
  }2019]{leng2019deepreviewer}
Youfang Leng, Li~Yu, and Jie Xiong.
\newblock 2019.
\newblock Deepreviewer: Collaborative grammar and innovation neural network for
  automatic paper review.
\newblock In {\em Proceedings of the ICMI}, pages 395--403.

\bibitem[\protect\citename{Li \bgroup et al.\egroup }2019]{li2019neural}
Siqing Li, Wayne~Xin Zhao, Eddy~Jing Yin, and Ji-Rong Wen.
\newblock 2019.
\newblock A neural citation count prediction model based on peer review text.
\newblock In {\em Proceedings of the EMNLP}, pages 4916--4926.

\bibitem[\protect\citename{Lin \bgroup et al.\egroup }2017]{lin2017structured}
Zhouhan Lin, Minwei Feng, Cicero Nogueira~dos Santos, Mo~Yu, Bing Xiang, Bowen
  Zhou, and Yoshua Bengio.
\newblock 2017.
\newblock A structured self-attentive sentence embedding.
\newblock In {\em Proceedings of the ICLR}.

\bibitem[\protect\citename{Pang and Lee}2005]{pang2005seeing}
Bo~Pang and Lillian Lee.
\newblock 2005.
\newblock Seeing stars: Exploiting class relationships for sentiment
  categorization with respect to rating scales.
\newblock In {\em Proceedings of the ACL}, pages 115--124.

\bibitem[\protect\citename{Peng \bgroup et al.\egroup }2018]{peng2018large}
Hao Peng, Jianxin Li, Yu~He, Yaopeng Liu, Mengjiao Bao, Lihong Wang, Yangqiu
  Song, and Qiang Yang.
\newblock 2018.
\newblock Large-scale hierarchical text classification with recursively
  regularized deep graph-cnn.
\newblock In {\em Proceedings of the WWW}, pages 1063--1072. International
  World Wide Web Conferences Steering Committee.

\bibitem[\protect\citename{Peng \bgroup et al.\egroup
  }2019]{peng2019hierarchical}
Hao Peng, Jianxin Li, Senzhang Wang, Lihong Wang, Qiran Gong, Renyu Yang,
  Bo~Li, Philip Yu, and Lifang He.
\newblock 2019.
\newblock Hierarchical taxonomy-aware and attentional graph capsule rcnns for
  large-scale multi-label text classification.
\newblock {\em TKDE}.

\bibitem[\protect\citename{Pennington \bgroup et al.\egroup
  }2014]{pennington2014glove}
Jeffrey Pennington, Richard Socher, and Christopher Manning.
\newblock 2014.
\newblock Glove: Global vectors for word representation.
\newblock In {\em Proceedings of the EMNLP}, pages 1532--1543.

\bibitem[\protect\citename{Qiao \bgroup et al.\egroup }2018]{qiao2018new}
Chao Qiao, Bo~Huang, Guocheng Niu, Daren Li, Daxiang Dong, Wei He, Dianhai Yu,
  and Hua Wu.
\newblock 2018.
\newblock A new method of region embedding for text classification.
\newblock In {\em Proceedings of the ICLR}.

\bibitem[\protect\citename{Qu \bgroup et al.\egroup }2010]{qu2010bag}
Lizhen Qu, Georgiana Ifrim, and Gerhard Weikum.
\newblock 2010.
\newblock The bag-of-opinions method for review rating prediction from sparse
  text patterns.
\newblock In {\em Proceedings of the COLING}, pages 913--921. Association for
  Computational Linguistics.

\bibitem[\protect\citename{Rush \bgroup et al.\egroup }2015]{rush2015neural}
Alexander~M Rush, Sumit Chopra, and Jason Weston.
\newblock 2015.
\newblock A neural attention model for abstractive sentence summarization.
\newblock In {\em Proceedings of the EMNLP}, pages 379--389.

\bibitem[\protect\citename{Shen \bgroup et al.\egroup }2018a]{shen2018bi}
T~Shen, T~Zhou, G~Long, J~Jiang, and C~Zhang.
\newblock 2018a.
\newblock Bi-directional block self-attention for fast and memory-efficient
  sequence modeling.
\newblock In {\em Proceedings of the ICLR}.

\bibitem[\protect\citename{Shen \bgroup et al.\egroup }2018b]{shen2018disan}
Tao Shen, Tianyi Zhou, Guodong Long, Jing Jiang, Shirui Pan, and Chengqi Zhang.
\newblock 2018b.
\newblock Disan: Directional self-attention network for rnn/cnn-free language
  understanding.
\newblock In {\em Proceedings of the AAAI}, pages 5446--5455.

\bibitem[\protect\citename{Vaswani \bgroup et al.\egroup
  }2017]{vaswani2017attention}
Ashish Vaswani, Noam Shazeer, Niki Parmar, Jakob Uszkoreit, Llion Jones,
  Aidan~N Gomez, {\L}ukasz Kaiser, and Illia Polosukhin.
\newblock 2017.
\newblock Attention is all you need.
\newblock In {\em Proceedings of the NIPS}, pages 5998--6008.

\bibitem[\protect\citename{Wang}2018]{wang2018disconnected}
Baoxin Wang.
\newblock 2018.
\newblock Disconnected recurrent neural networks for text categorization.
\newblock In {\em Proceedings of the ACL}, pages 2311--2320.

\bibitem[\protect\citename{Yang \bgroup et al.\egroup
  }2016]{yang2016hierarchical}
Zichao Yang, Diyi Yang, Chris Dyer, Xiaodong He, Alex Smola, and Eduard Hovy.
\newblock 2016.
\newblock Hierarchical attention networks for document classification.
\newblock In {\em Proceedings of the NAACL}, pages 1480--1489.

\bibitem[\protect\citename{Yang \bgroup et al.\egroup }2018]{yang2018automatic}
Pengcheng Yang, Xu~Sun, Wei Li, and Shuming Ma.
\newblock 2018.
\newblock Automatic academic paper rating based on modularized hierarchical
  convolutional neural network.
\newblock In {\em Proceedings of the ACL}, pages 496--502.

\bibitem[\protect\citename{Yin and Sch{\"u}tze}2018]{yin2018attentive}
Wenpeng Yin and Hinrich Sch{\"u}tze.
\newblock 2018.
\newblock Attentive convolution: Equipping cnns with rnn-style attention
  mechanisms.
\newblock {\em TACL}, 6:687--702.

\bibitem[\protect\citename{Zhang \bgroup et al.\egroup
  }2010]{zhang2010comparison}
DongMei Zhang, Shengen Li, Cuiling Zhu, Xiaofei Niu, and Ling Song.
\newblock 2010.
\newblock A comparison study of multi-class sentiment classification for
  chinese reviews.
\newblock In {\em Proceedings of the FSK}, volume~5, pages 2433--2436. IEEE.

\end{thebibliography}

\section*{Appendix}
\subsection*{Additional Details of Experimental Setting}
All the experiments are conducted on GPU devices. 
The software platforms are Python 3.6.8 and Tensorflow 1.13.1. 
\subsection*{Additional Experimental Results}
For the review rating prediction task, we provide additional experimental results on OpenReview dataset under Macro-F1 and Micro-F1 metrics. The results are shown in Table \ref{table:results_additional_10cls}. Our proposed framework still achieves the best results (in bold) on Macro-F1 and Micro-F1, it outperforms all the baselines in a similar trend as under other metrics such as Accuracy, DM and OP.
\begin{table}
\small
\centering
\begin{tabular}{c|c| c c} 
 \hline
		\multirow{3}{*}{}  & \multirow{3}{*}{Models}    
		& \multicolumn{2}{c}{Review Rating Prediction} \\ \cline{3-4} 
		\multirow{3}{*}{}  & \multirow{3}{*}{}
		& \multicolumn{2}{c}{OpenReview} \\ \cline{3-4}
		& & Macro-F1 & Micro-F1 \\ [0.5ex] 
 \hline
 \multirow{10}{0.4cm}{\rotatebox{90}{\tabincell{c}{Flat Baselines}}}
 & NB & 0.037 & 0.083 \\ [0.5ex]
 & LR & 0.138 & 0.280 \\ [0.5ex]
 & SVM & 0.142 & 0.283 \\ [0.5ex]
 \cline{2-4}
 & RNN \cite{bahdanau2014neural} & 0.109 & 0.219 \\ [0.5ex]
 & TextCNN \cite{kim2014convolutional} & 0.121 & 0.245 \\ [0.5ex]
 & TextRCNN \cite{lai2015recurrent} & 0.134 & 0.250 \\ [0.5ex]
 & VDCNN \cite{conneau2016very} & 0.102 & 0.193 \\ [0.5ex]
 & DPCNN \cite{johnson2017deep} & 0.137 & 0.262 \\ [0.5ex]
  \cline{2-4}
 & Transformer \cite{vaswani2017attention} & 0.040 & 0.085 \\ [0.5ex]
 & SA-Sent-EM \cite{lin2017structured} & 0.182 & 0.310 \\ [0.5ex]
 \hline
 \multirow{4}{0.4cm}{\rotatebox{90}{\tabincell{c}{Hierarchical\\ Baselines}}}
 & HAN \cite{yang2016hierarchical}-extended & 0.167 & 0.311 \\ [0.5ex]
 \cline{2-4}
 & Bert-base \cite{devlin2018bert} & 0.188 & 0.297 \\ [0.5ex]
 & Bert-large \cite{devlin2018bert} & 0.182 & 0.300 \\ [0.5ex]
 & SciBert \cite{beltagy2019scibert} & 0.195 & 0.311 \\ [0.5ex]
 \hline
 \multirow{4}{0.4cm}{\rotatebox{90}{\tabincell{c}{Ours}}}
 & {\methodname} (Glove) & \textbf{0.205} & \textbf{0.338} \\ [0.5ex]
 & {\methodname} (Bert-base) & \textbf{0.223} & \textbf{0.330} \\ [0.5ex]
 & {\methodname} (SciBert) & \textbf{0.216} & \textbf{0.325} \\ [0.5ex]
 & {\methodname} (Bert-large) & \textbf{0.226} & \textbf{0.339} \\ [0.5ex]
\hline
\end{tabular}
\caption{Additional results of review rating prediction on OpenReview dataset under Macro-F1 and Micro-F1 metrics.}
\label{table:results_additional_10cls}
\end{table}

\end{document}